\pdfoutput = 1
\documentclass[twoside,11pt]{article}
\usepackage[preprint]{jmlr2e}

\usepackage[utf8]{inputenc} 
\usepackage[T1]{fontenc}    
\usepackage{url}            
\usepackage{booktabs}       
\usepackage{amsfonts}       
\usepackage{nicefrac}       
\usepackage{microtype}      
\usepackage{xcolor}         
\usepackage{pifont}         
\usepackage{breakcites}
\usepackage{listings}
\usepackage{array}
\newcommand{\PreserveBackslash}[1]{\let\temp=\\#1\let\\=\temp}
\newcolumntype{C}[1]{>{\PreserveBackslash\centering}p{#1}}
\newcolumntype{R}[1]{>{\PreserveBackslash\raggedleft}p{#1}}
\newcolumntype{L}[1]{>{\PreserveBackslash\raggedright}p{#1}}

\hypersetup{
    colorlinks=true,
    allcolors=brown
}
\usepackage{physics} 
\usepackage[linesnumbered,ruled,vlined]{algorithm2e}
\usepackage{xcolor}
\usepackage{amsmath}
\usepackage{amssymb}
\usepackage{numprint}
\usepackage{dsfont}
\usepackage{bm}
\usepackage{bbm}
\usepackage{float}
\usepackage{sectsty}
\usepackage{graphicx} 
\usepackage{subcaption}
\usepackage{enumitem}
\usepackage{authblk}
\usepackage{multido}

\definecolor{delim}{RGB}{20,105,176}
\definecolor{numb}{RGB}{106, 109, 32}
\definecolor{string}{rgb}{0.64,0.08,0.08}

\lstdefinelanguage{json}{
    numbers=left,
    numberstyle=\small,
    frame=single,
    rulecolor=\color{black},
    showspaces=false,
    showtabs=false,
    breaklines=true,
    postbreak=\raisebox{0ex}[0ex][0ex]{\ensuremath{\color{gray}\hookrightarrow\space}},
    breakatwhitespace=true,
    basicstyle=\ttfamily\small,
    upquote=true,
    morestring=[b]",
    stringstyle=\color{string},
    literate=
     *{0}{{{\color{numb}0}}}{1}
      {1}{{{\color{numb}1}}}{1}
      {2}{{{\color{numb}2}}}{1}
      {3}{{{\color{numb}3}}}{1}
      {4}{{{\color{numb}4}}}{1}
      {5}{{{\color{numb}5}}}{1}
      {6}{{{\color{numb}6}}}{1}
      {7}{{{\color{numb}7}}}{1}
      {8}{{{\color{numb}8}}}{1}
      {9}{{{\color{numb}9}}}{1}
      {\{}{{{\color{delim}{\{}}}}{1}
      {\}}{{{\color{delim}{\}}}}}{1}
      {[}{{{\color{delim}{[}}}}{1}
      {]}{{{\color{delim}{]}}}}{1},
}

\title{PyRCA: A Library for Metric-based Root Cause Analysis}
    \author{
    \centering
    Chenghao Liu\thanks{Equal contribution}$^*$, Wenzhuo Yang$^*$, Himanshu Mittal, Manpreet Singh, \\Doyen Sahoo\thanks{Correspondence: \texttt{\{dsahoo, shoi\}@salesforce.com}}$^\dagger$, Steven C. H. Hoi$^\dagger$\\Salesforce AI\\
    \url{https://github.com/salesforce/PyRCA}
    }
\affil{\centering}%

\begin{document}
\maketitle

\begin{abstract}
We introduce PyRCA,
an open-source Python machine learning library of Root Cause Analysis (RCA) for Artificial Intelligence for IT Operations (AIOps). It provides a holistic framework to uncover the complicated metric causal dependencies and automatically locate root causes of incidents. It offers a unified interface for multiple commonly used RCA models, encompassing both graph construction and scoring tasks. This library aims to provide IT operations staff, data scientists, and researchers a one-step solution to rapid model development, model evaluation and deployment to online applications. In particular, our library includes various causal discovery methods to support causal graph construction, and multiple types of root cause scoring methods inspired by Bayesian analysis, graph analysis and causal analysis, etc. Our GUI dashboard offers practitioners an intuitive point-and-click interface, empowering them to easily inject expert knowledge through human interaction. With the ability to visualize causal graphs and the root cause of incidents, practitioners can quickly gain insights and improve their workflow efficiency. This technical report introduces PyRCA's architecture and major functionalities, while also presenting benchmark performance numbers in comparison to various baseline models. Additionally, we demonstrate PyRCA's capabilities through several example use cases.
\end{abstract}

\begin{keywords}
root cause analysis, causal analysis, time series, anomaly detection, machine learning, AIOps, Python, scientific toolkit
\end{keywords}

\section{Introduction}
Root Cause Analysis (RCA) is a commonly used method for ensuring the reliability and efficiency of production systems in various domains, including IT operations ~\citep{soldani2022anomaly}, telecommunications ~\citep{zhang2021influence}, and epidemiology ~\citep{landsittel2020narrative}. Incidents that indicate the actual inability of a module to perform its function in the system are inevitable in practice due to the complexity of modern systems. RCA involves identifying the underlying reasons for an observed incident, with the ultimate goal of providing possible explanations. By examining various factors and determining their causal relationships, RCA can improve diagnosis and triage performance, and thus reduce the negative influence of the incidents. 

As more Internet applications are being deployed on the cloud, ensuring the quality of cloud systems and user experience has become increasingly important. Incidents in these systems can lead to poor user experience and significant economic loss. One way to address this is by building a monitoring system that collects and tracks Key Performance Indicators (KPI) from running applications. Anomalies in these KPI metrics can be treated as incidents. When an incident occurs, engineers typically collect all related metrics and investigate their behaviors to identify the root cause or clues for further diagnosis. For example, in an e-commerce system as shown in Figure \ref{fig:example}, if the response time of the web service increases significantly, engineers would check the response time of upstream services in the causal graph. If the response time of a database service is high, it may indicate that the long response time of the web service is caused by the database service, and the incident can be mitigated by restarting the database server. However, modern cloud systems often comprise a large number of components connected through complex dependencies, running in a distributed environment. With thousands or more KPI metrics to explore for each incident, manually checking so many potentially relevant metrics each day is time-consuming, labor-intensive and error-prone for engineers. Therefore, an automated RCA toolbox is highly desirable.

We are motivated to develop a holistic PyRCA solution, a Python library for root cause analysis, to meet the needs of both industrial and academic use cases. PyRCA is the first open source RCA library that offers an end-to-end framework that includes data loading, causal graph discovery, root causes localization and RCA results visualization. PyRCA supports multiple causal graph construction and root cause scoring models. Additionally, it comes with a GUI dashboard to conduct RCA in an interactive way, which better aligns with the user experience in real-world scenarios. PyRCA's key features are:

\begin{figure}[t]
    \centering
    \includegraphics[width=1\columnwidth]{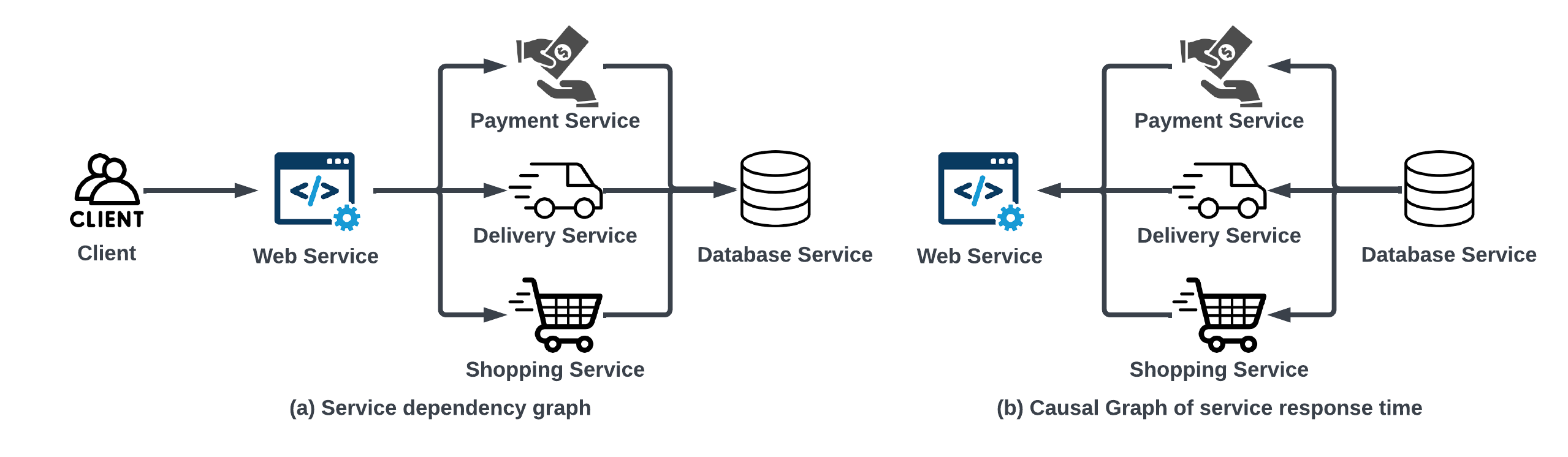}
    \caption{An example of e-commerce system, which includes web service, payment service, delivery service, shopping service and database service. \label{fig:example} Figure (a) shows the dependencies between each service. Figure (b) shows the causal graph of response time of each services by inverting the dependency graph.}
\end{figure}

\begin{itemize}
    \item PyRCA offers a standardized and highly adaptable framework for loading metric data with the commonly used \texttt{pandas.DataFrame} format and benchmarking a diverse set of RCA models.
    \item It offers a collection of varied models for both uncovering causal graph and identifying root causes, all accessible through a unified interface. Additionally, advanced users have the option to fully configure each model to meet their specific needs. Models include PC ~\citep{spirtes2000causation}, GES~\citep{chickering2002optimal}, random walk ~\citep{wang2018cloudranger}, hypothesis-testing ~\citep{li2022causal}, etc. 
    \item The RCA models provided in the library can be enhanced by incorporating user-provided domain knowledge, making them more robust when dealing with noisy metric data.
    \item Developers can easily extend PyRCA by implementing a single class inherited from the RCA base class to add new RCA models.
    \item The PyRCA library also offers a visualization tool that allows users to easily incorporate domain knowledge, examine RCA results, and compare different models, all without requiring any coding. 
\end{itemize}

\subsection{Related Work}
While RCA is a critical machine learning problem that is in high demand with diverse applications, there is current no well-developed open source tool available to address this need with a one-stop solution. The most relevant open source library in the field of RCA is CIRCA \protect\footnote{\url{https://github.com/netmanaiops/circa}}, which mainly focus on root cause scoring methods. It  contains various baselines and simulation data generation for the hypothesis-testing method \citep{li2022causal}. For the graph construction problem, several open source libraries supporting causal discovery algorithms can handle it, such as Tetrad \protect\footnote{\url{https://github.com/cmu-phil/tetrad}}, causalai\protect\footnote{\url{https://github.com/salesforce/causalai/tree/main/causalai}}, and causal-learn\protect\footnote{\url{https://github.com/py-why/causal-learn}}.

\pdfoutput = 1
\section{RCA Framework and Design Principles}

\begin{figure}[t]
    \centering
    \includegraphics[width=1\columnwidth]{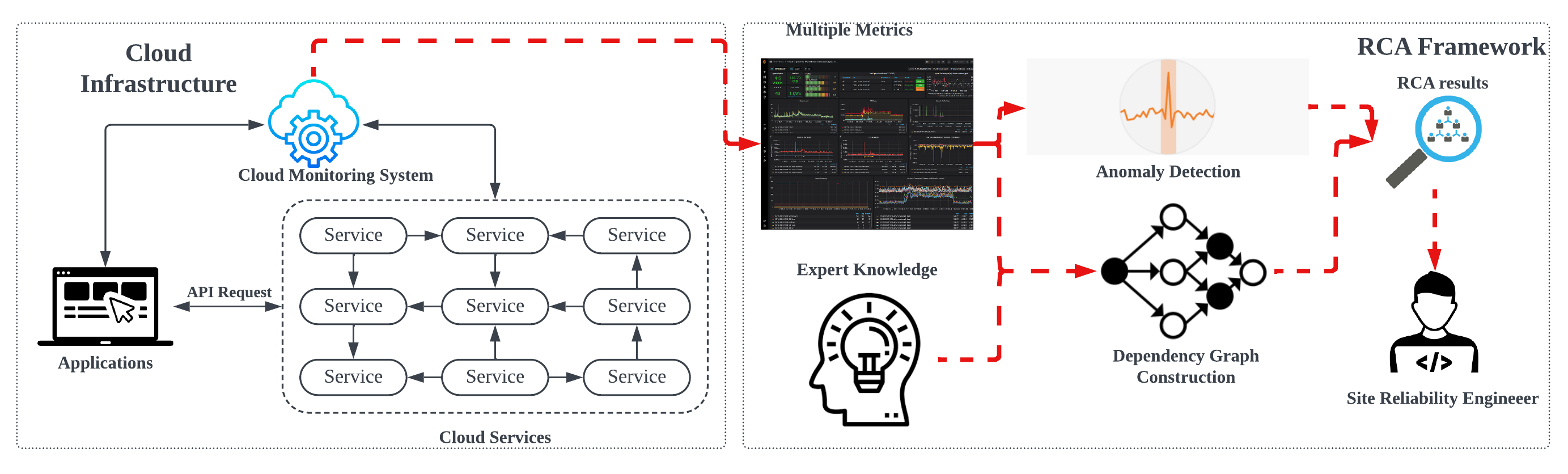}
    \caption{The framework of RCA. Left figure demonstrates a production system which contains large-number of service with complex interdependencies. The cloud monitoring system collects the details of each service request and status in a streaming way. Right figure demonstrates the pipeline of RCA. 
    \label{fig:framework}}
\end{figure}
We first show an example how RCA works when integrating with a production system in Figure (\ref{fig:framework}). A typical production system is comprised of a large number of services with complex inter-dependencies. To ensure the reliability of the entire system, a monitoring system is integrated to periodically collect various measures that monitor the health of each service. When anomaly metrics are detected by the anomaly detection module, they usually indicate the corresponding service failure, which is not functional and seriously impacts user experience. The anomaly detection module then automatically triggers the root cause localization task for the RCA module. Meanwhile, the RCA module leverages multiple metrics from the production system as well as expert knowledge to construct a causal graph. By considering the anomaly metrics and their dependencies in the graph, the RCA module is able to calculate root cause scores and present the results to site reliability engineers to assist them in subsequent remediation actions. Given this, the RCA objective can be formulated as follows: Given the anomaly metrics, RCA will try to localize the top-K metrics that are most likely to be the root cause of the anomaly metrics \citep{meng2020localizing}.

The key design principle of PyRCA is to provide a simple but unified framework for RCA problems. PyRCA allows users to apply multiple RCA models and visualize the corresponding RCA results simultaneously. This aims to make the PyRCA library easy to customize, allowing users to develop a customized RCA workflow by simply modifying the configuration file. Additionally, the library is easy to extend, enabling users to add new causal graph construction and root cause scoring methods without affecting the library framework. Finally, PyRCA is easy to use, supporting an interactive dashboard to incorporate expert knowledge and demonstrate results. 

At a high level, PyRCA library API is composed of three main components - the input layer directly loads metric data in \texttt{pandas.DataFrame} format and parses expert knowledge from configuration file with \texttt{YAML} format, the model layer implements a wide range of model for anomaly detection, causal graph construction and root cause scoring methods. The output layer supports causal graph and root cause analysis results visualization and evaluation. Additionally, it includes a data simulation tool for empirical analysis

\begin{figure}[t]
    \centering
    \includegraphics[height=0.4\columnwidth, width=0.8\columnwidth]{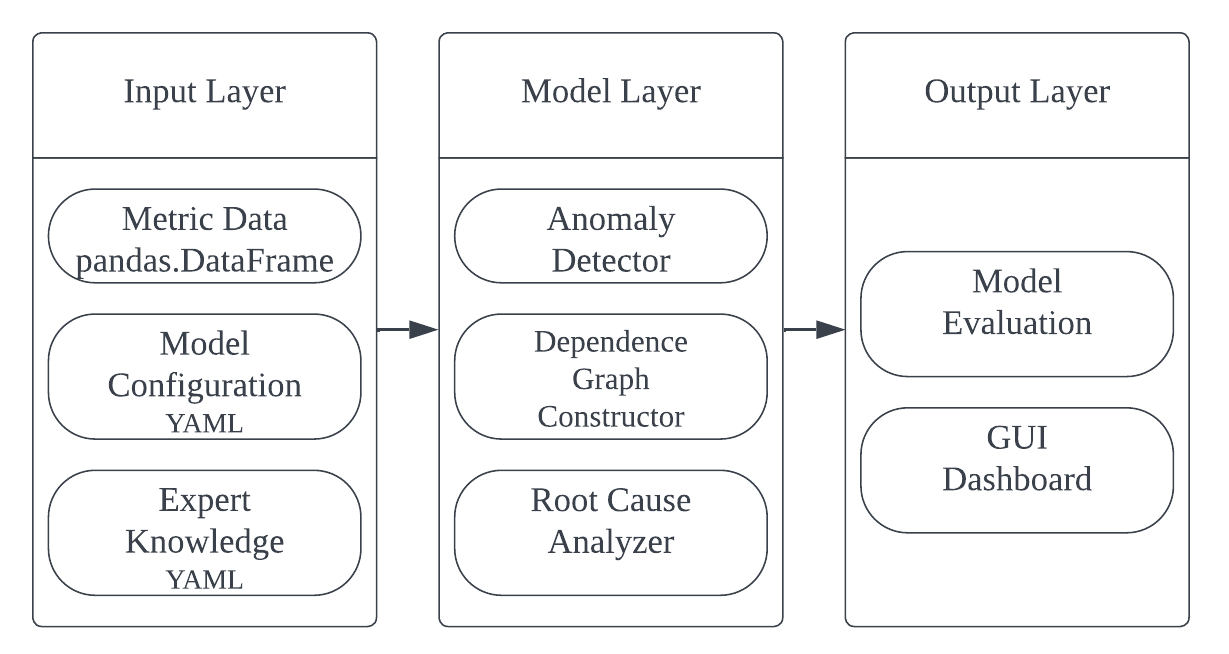}
    \caption{The main architecture of PyRCA. \label{fig:arch}}
\end{figure}

\subsection{Input Layer} PyRCA loads metric data in \texttt{pandas.DataFrame} format, where each column represents a specific metric. This format is general to support both tabular data and time series data, which uses timestamp as the index. The expert knowledge file has a \texttt{YAML} format. It supports to add constraints for forbidding a connection, enforcing a connection, declaring the root nodes and leaf nodes. We demonstrate an example of configuration file in the following:

\begin{lstlisting}[language=json]
causal-graph:
  root-nodes: ["A", "B"] # Metrics A and B must the root nodes
  leaf-nodes: ["E", "F"] # Metrics E and F must be the leaf nodes
  forbids:
    - ["A", "E"] # There is no connection from metric A to metric E,
  requires: 
    - ["A", "C"] # There is a connection from metric A to metric C
\end{lstlisting}

\subsection{Model Layer} Flexibility in model selection is crucial, as no single model can perform well across all use cases. Therefore, it is important to offer a broad range of models for users to choose from. To provide users with a seamless and transparent experience in accessing different modeling options, we have unified all PyRCA models under three categories: anomaly detection, causal graph construction, and root cause scoring. This makes it easier for users to explore and utilize the different models without having to navigate through various APIs or interfaces. All PyRCA models can be easily accessed and utilized through two common APIs. Users can initialize any model by passing a config object that includes model specific hyperparameters, and can then use the \texttt{model.train} method to train the model on the desired data. This unified approach allows for greater ease of use and flexibility in the selection and customization of models. For instance, if the developer want to contribute a customized causal graph construction model, he is required to create the configuration class that inherits from \texttt{pyrca.base.CausalModelConfig}. Then, he need to create the model class that inherits from \texttt{pyrca.analyzers.base.CausalModel}. The constructor for the new model takes the new configuration instance as its input. Finally, he need to implement the \texttt{\_train} function that returns the discovered causal graph. It should be noted that the anomaly detection model need to implement an additional function \texttt{\_predict} that returns the anomaly scores, and root cause scoring model need to implement an additional function \texttt{find\_root\_causes} that returns \texttt{pyrca.analyzers.base.RCAResults} instance storing root cause analysis results.

\subsection{Output Layer} \label{sec:outputlayer} PyRCA provide an interactive dashboard to run customized RCA models and display the RCA results along with the estimated causal graph.  The users can access this tool by running the command \texttt{python -m pyrca.tools}.  The dashboard offers a range of features, such as experimenting with various RCA models, adjusting the model's hyperparameters, imposing expert knowledge constraints like root/leaf nodes or forbidden/required links, and visualizing the generated causal graph. This makes it convenient for users to manually update causal graphs with expert knowledge. Once satisfied with the results, the users can download the outcomes to be employed by the RCA methods supported in PyRCA. Since the RCA pipeline includes two main steps: causal graph construction and root cause scoring. Therefore, we can evaluate their performance, respectively. To measure the performance of the estimated graph and truth graph, we can compute the precision and recall of the adjacent confusion matrix and the Structural Hamming Distance (SHD), which is computed as wrongly directed edges.
To measure the performance of the root cause localization, we can calculate the recall with the top-k results, which is widely used in existing works \citep{meng2020localizing, li2022causal, yu2021microrank}. These related evaluation metrics are implemented in \texttt{pyrca.utils.evalution}.

In addition, PyRCA also includes an example application that uses the \texttt{BayesianNetwork} for root cause scoring. The \texttt{config} folder contains the setups for the stats-based anomaly detector and the domain knowledge, while the \texttt{models} folder stores the causal graph and the trained Bayesian network. The \texttt{RCAEngine} class in the \texttt{rca.py} file provides all the necessary methods for building causal graphs, training Bayesian networks, and finding root causes using the modules provided by PyRCA.

\subsection{Simulated Data Generation} 
PyRCA offers a tool for simulated data generation. The generation process is as follows:
i). run \texttt{DAGGen.gen} method to generate the Directed Acyclic Graph (DAG), given the number of nodes and the number of edges. ii). Given the DAG and specification of Structural Equation Model (e.g. noise form, function form.), run \texttt{DataGen.gen} to generate normal data/ 3). Given the normal data generation process and the way of injecting incident, generate the abnormal data by running \texttt{AnomalyDataGen.gen}.

\pdfoutput = 1
\section{Root Cause Analysis Model}
\begin{figure}[t]
    \centering
    \includegraphics[width=1\columnwidth]{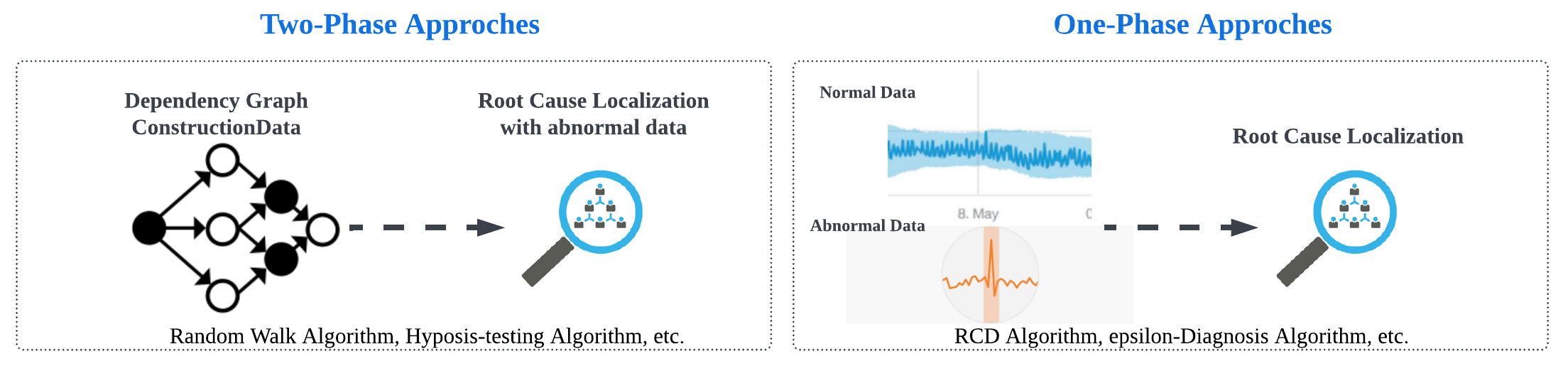}
    \caption{The taxonomy of RCA models. \label{fig:tax}}
\end{figure}
\subsection{Two-Phase RCA Models}
\label{sec:two-phase}
One common type of RCA models involves two steps. The first step is constructing causal graphs based on observed metrics and domain knowledge. The second step is extracting anomalous subgraphs or paths based on observed anomalies. Usually, 
these causal graphs can be reconstructed from the topology of a specific application (i.e., domain knowledge from log analysis and trace analysis). 
In cases where the service or call graphs are not available or only partially available, constructing the topology graph of the production system can be challenging. Fortunately, with the observed metrics, causal discovery models can play a significant role in constructing the causal graph that describes the causal relationships between these metrics in a data driven way. This is particularly useful when investigating the relationships between the monitored metrics instead of API calls. The most popular causal discovery algorithm
applied in RCA is the well-known PC algorithm \citep{spirtes2000causation}, GES algorithm \citep{chickering2002optimal}. Once the causal graph has been constructed, there are several methods that can be used to identify the possible root causes of observed anomalies. Two commonly used methods are the random walk algorithm \citep{yu2021microrank} and hypothesis-testing algorithms \citep{li2022causal}. In the following we will provide a detailed introduction of the several algorithms mentioned above.

\subsubsection{PC Algorithm} 
The PC algorithm is a well-known constraint-based causal discovery algorithm that relies on the i.i.d. sampling assumption and the no latent confounders assumption. It provides a search architecture that can be combined with many statistical procedures to determine conditional independencies. Specifically, it starts with a complete undirected graph and recursively eliminate edges between variables that are conditionally independent. Once the edges have been eliminated, the algorithm will only test the remaining edges in the next iteration. This key feature makes the PC algorithm efficient in sparse graph. The orientations of the edges are then determined by finding V-structures followed
by orientation propagation \citep{spirtes2000causation}. 

\subsubsection{GES Algorithm} 
The GES algorithm, a typical score-based causal discovery algorithm, is introduced in \citep{chickering2002optimal}. The key idea is to assign a score to each candidate causal graph for measuring how well the candidate causal graph fits dataset. Unlike the PC algorithm, which starts with a complete undirected graph, GES begins with a graph without any edges, signifying that all variables are independent of each other. GES then greedily adds needed edges one at a time in the orientation that minimizes the Bayes Information Score (BIC), which denotes the likelihood penalized for complexity to reduce overfitting. When the score can no longer be decreased, GES removes edges one at a time, as long as it decreases the BIC until no further edges can be removed.

\subsubsection{Random Walk Algorithm} 
After discovering the causal graph, possible root causes of anomaly metrics can be determined through the random walk algorithm \citep{wang2018cloudranger}. A random walk is a stochastic process that moves randomly from one node to another in a graph. It is guided by a transition probability matrix that specifies the likelihood of moving from one node to another. In the context of RCA, the random walk algorithm is based on the assumption that a metric that is highly correlated with the anomaly metrics is more likely to be the root cause. By conducting random walks on the graph, the RCA model can calculate root cause scores for each metric and identify the top-K metrics that are most likely to be the candidate root cause of the anomaly metrics. The transition probability matrix is a crucial component of random walk approaches for RCA. It determines the likelihood of moving from one node to another during a random walk. Typically, there are three steps involved in computing the transition probability matrix \citep{cheng2023ai}. The first is the forward step, which calculates the probability of walking from a node to one of its parents. The second is the backward step, which calculates the probability of walking from a node to one of its children. The third is the self step, which calculates the probability of staying at the current node. These probabilities can be estimated based on the graph structure, expert knowledge, or statistical analysis of the available data .

\subsubsection{Hypothesis-Testing Algorithm} 
Another common algorithm for root cause scoring is hypothesis testing \citep{li2022causal}. The key idea of this approach is to map anomalies to unexpected interventions. Specifically, it assumes that the normal data come from the observational distribution, while the abnormal data come from an intervention that has disrupted the system. The goal of root cause scoring using hypothesis testing is to recognize the intervention by identifying the variables (nodes) that have been intervened upon, given the normal and abnormal data. This can be achieved using statistical tests to compare the observed variable distribution in abnormal data with the estimated variable distribution obtained by training the estimated model with the normal data. If the difference is statistically significant, the corresponding variable (node) is a potential root cause.

\subsection{One-Phase RCA Models}
Although two-phase RCA models offer powerful explainability by constructing a causal graph, the runtime of causal graph construction algorithms can be a limiting factor. In the worst-case scenario, the runtime can be exponential in the number of variables (nodes), which can hinder their application in real-world scenarios. In contrast, one-phase RCA models directly handle normal and abnormal data to output the root causes and have the ability to efficiently handle thousands or even millions of metrics.

\subsubsection{$\epsilon$-Diagnosis Algorithm} 
The $\epsilon$-Diagnosis algorithm \citep{shan2019diagnosis} is a low-cost RCA model that can detect the root causes of small window long-tail latency for web services. The algorithm assumes that the root cause metrics of an abnormal service have significantly changed between the abnormal and normal periods. To identify the root causes, the $\epsilon$-Diagnosis Algorithm uses the two-sample test algorithm and $\epsilon$-statistics for measuring the similarity of time series. In the two-sample test, one sample (normal sample) is drawn from the snapshot during the normal period, while the other sample (anomaly sample) is drawn during the anomalous period. If the difference is statistically significant, the corresponding metrics of the samples are potential root causes.

\subsubsection{RCD Algorithm} 
The RCD algorithm \citep{ikram2022root} is another low-cost RCA model that can uncover the dependencies between metrics using both normal and abnormal data. Its key features include the ability to model the anomaly as an intervention on the root cause node and the introduction of a binary indicator variable for normal and abnormal data. Compared to two-stage RCA models, RCD does not need to learn the full causal graph for RCA, allowing it to handle a larger set of metrics.

\pdfoutput = 1
\section{Experiments}
In this section, we compare the performance of different RCA models on the simulated data and then demonstrate how to use the dashboard.

\subsection{Data Generation}

We first describe the proposed data generation process. The generation pipeline follows five steps: (i) specify and generate the Directed Acyclic Graph (DAG), (ii) specify and generate a Structural Causal Model \citep{pearl2000models} (including function form and noise form), (iii) generate synthetic data, (iv) specify the root cause nodes and how they change the corresponding SCM, and (v) generate abnormal data. Specifically, we generate a simulated dataset with 20 nodes and 30 edges. The root cause nodes and the change mechanisms that affect the DAG are randomly selected. We generate 500 graphs and 5000 samples per graph. Evaluation metrics averaged over the 500 graphs will be presented. As introduced in Section \ref{sec:outputlayer}, we report Recall@k to measure the performance of RCA models.

\begin{table}[t]
    \centering
    \begin{tabular}{cccc}
    \hline
         & Recall@1 & Recall@3 & Recall@5 \\ \hline\hline
        $\epsilon$-Diagnosis & 0.06 $\pm$ 0.02 & 0.16 $\pm$ 0.04 & 0.16 $\pm$ 0.04 \\ 
        RCD & 0.28 $\pm$ 0.05 & 0.29 $\pm$ 0.05 & 0.30 $\pm$ 0.05 \\ 
        Local-RCD & 0.44 $\pm$ 0.05 & 0.70 $\pm$ 0.05 & 0.70 $\pm$ 0.05 \\ 
        \hline
        RW & 0.07 $\pm$ 0.03 & 0.20 $\pm$ 0.04 & 0.24 $\pm$ 0.04 \\ 
        RW-pc & 0.06 $\pm$ 0.02 & 0.17 $\pm$ 0.04 & 0.21 $\pm$ 0.04 \\ 
        BI & 0.15 $\pm$ 0.04 & 0.35 $\pm$ 0.05 & 0.43 $\pm$ 0.05 \\ 
        BI-pc & 0.11 $\pm$ 0.03 & 0.30 $\pm$ 0.05 & 0.40 $\pm$ 0.05 \\ 
        HT & \textbf{1.00 $\pm$ 0.00} & \textbf{1.00 $\pm$ 0.00} & \textbf{1.00 $\pm$ 0.00} \\ 
        HT-pc & 0.95 $\pm$ 0.02 & \textbf{1.00 $\pm$ 0.00} & \textbf{1.00 $\pm$ 0.00} \\ 
        HT-adj & \textbf{1.00 $\pm$ 0.00} & \textbf{1.00 $\pm$ 0.00}& \textbf{1.00 $\pm$ 0.00} \\ 
        HT-adj-pc & 0.77 $\pm$ 0.04 & 0.92 $\pm$ 0.03 & 0.92 $\pm$ 0.03 \\ \hline
    \end{tabular}
    \caption{The RCA benchmark on the simulated dataset.  Best results are highlighted in bold. $\epsilon$-Diagnosis \citep{shan2019diagnosis} and RCD are one-phase RCA models, while the rest models are two-phase RCA models. Local-RCD denotes the RCD algorithm with localized learning \citep{ikram2022root}. RW denotes the random walk algorithm. BI denotes the Bayesian Inference algorithm which computes the root cause scores by estimating each SCM \citep{koyyalummal2022self}. HT denotes
    the hypothesis-testing algorithm. 
    HT-adj denotes the HT algorithm with descendant adjustment \citep{li2022causal}. For the two-phase models, the algorithms without suffix indicate that the root cause localization algorithm use the true causal graph for model training. The algorithms with suffix "pc" indicate the causal graph is estimated via PC algorithm.
     \label{tab:benchmark}}.  
\end{table}

\begin{table}[t]
    \centering
    \begin{tabular}{ccccc}
    \hline
         & Precision & Recall & F1 & SHD \\ \hline
        PC & \textbf{0.88 $\pm $ 0.06} & \textbf{0.71 $\pm$  0.09} & \textbf{0.78 $\pm $ 0.07} & \textbf{11.45 $\pm$ 3.66} \\ 
        GES & 0.46 $\pm $ 0.15 & 0.44 $\pm$  0.14 & 0.45 $\pm $ 0.14 & 32.53 $\pm$ 9.15 \\ \hline
    \end{tabular}
    \caption{The performance of causal graph construction between PC and GES algorithm \label{tab:graph}}
\end{table}

\subsection{Performance Comparison}
Table (\ref{tab:benchmark}) summarizes the RCA performance of different methods on the simulated dataset. It is evident that HT outperforms other baselines by a significant margin. Additionally, there is a notable difference in the performance of two-phase models when using a true graph versus an estimated graph, emphasizing the significance of constructing an accurate causal graph. As discussed in Section \ref{sec:two-phase}, the efficacy of using a causal discovery algorithm is heavily dependent on the dataset satisfying certain assumptions, such as linearity, latent confounders, etc. As a supplement, PyRCA supports an efficient way to incorporate expert knowledge, which can improve the quality of the uncovered causal graph. Although RCD and local-RCD utilize both normal and abnormal data to uncover metric dependencies, they do not perform as well as HT. This is because RCD is restricted to discrete variables, and the data discretization step can interfere with the independence analysis. In table (\ref{tab:graph}), we further compare the performance of PC algorithm and GES algorithm when constructing causal graph. We can observe that PC outpeforms GES in this simulated dataset.

\subsection{Dashboard Showcase}

\begin{figure}[th]
    \centering
    \includegraphics[width=1\columnwidth]{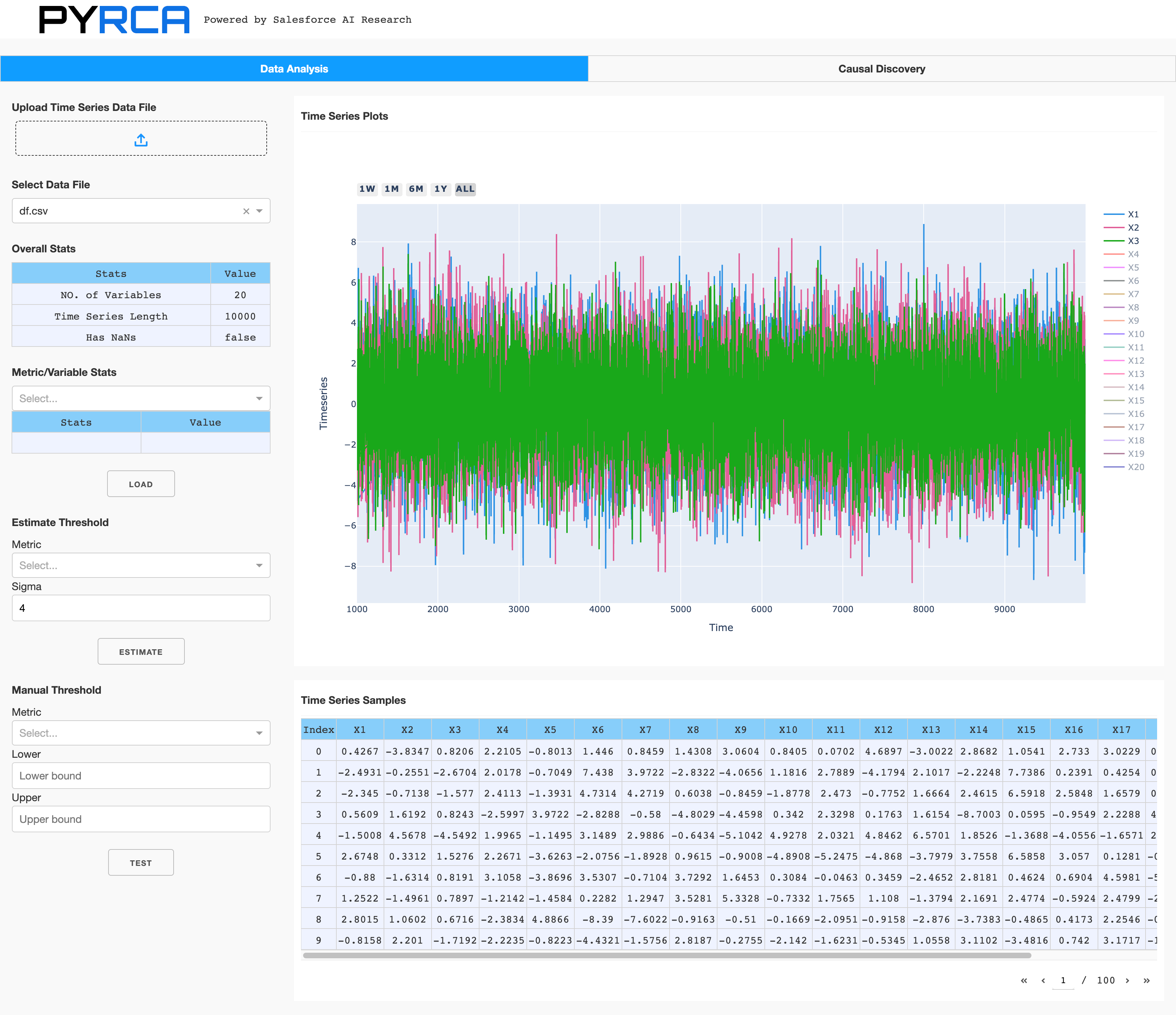}
    \caption{The interactive dashboard of PyRCA (Data Analysis Tab)\label{fig:gui}}
\end{figure}
\begin{figure}[th]
    \centering
    \includegraphics[width=1\columnwidth]{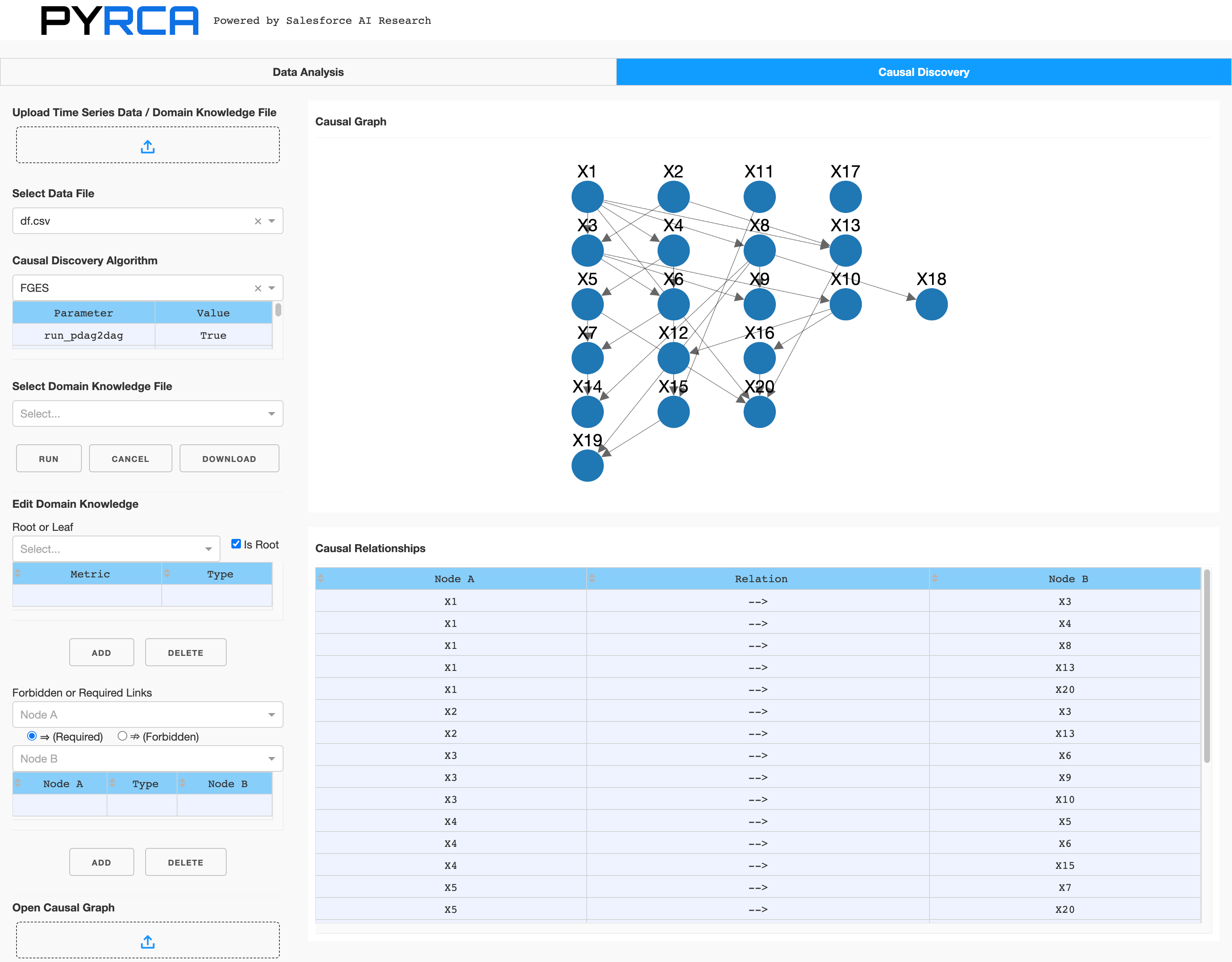}
    \caption{The interactive dashboard of PyRCA (Causal Graph Discovery Tab)\label{fig:gui_2}}
\end{figure}

Users can launch the dashboard app for data analysis and causal graph analysis by running \texttt{python -m pyrca.tools}. As shown in Figure (\ref{fig:gui}), the "Data Analysis`` tab enables users to upload metric data, visualize all the metrics, check basic statistics such as means and variances, and adjust the hyperparameters for stats-threshold based anomaly detectors. PyRCA supports a basic stats-based anomaly detector, \texttt{pyrca.outliers.stats}, which can be used to detect anomalous spikes in the data. However, if this anomaly detector is not suitable for your use case, you can explore other anomaly detectors offered by Merlion\protect\footnote{\url{https://github.com/salesforce/Merlion}}. Note that the time series data should be in CSV format, where the first column is the timestamp and the other columns represent the metrics. The "Causal Graph Discovery`` tab, as shown in Figure (\ref{fig:gui_2}), is used to construct causal graphs estimated from metric data. Users first need to upload the metric data, and the domain knowledge file (optional, in the YAML format). Then, users can choose the uploaded metric data to build the graph that describes the dependency relationships between different metrics. After setting the hyperparameters of the causal graph construction algorithm and the domain knowledge file path, users can click the "Run" button to generate the first version of the causal graph. Then, users can manually check for any missing or incorrect links in the generated graph. If the generated causal graph has errors, users can add additional constraints, such as root/leaf nodes and required/forbidden links, in the "Edit Domain Knowledge" card. After the new constraints are added, users can click the "Run" button again to refine the causal graph. If the generated causal graph is satisfactory, users can click the "Download" button to save the graph for future RCA model deployment.

In real-world applications, causal discovery methods often struggle to produce accurate causal graphs due to data issues. This app offers a user-friendly interface to allow interactive editing and revision of causal graphs.

\section{Conclusion and Future Work}

We introduce PyRCA, an open-source machine-learning library for Root Cause Analysis (RCA). PyRCA is designed to tackle many of the pain points associated with current industry workflows for IT operations. It provides unified, easily extensible interfaces and implementations for a wide range of RCA models. These features are combined in a flexible pipeline that evaluates the performance of a model quantitatively, along with a visualization module for more qualitative analysis. 

We are dedicated to continually improving PyRCA and plan to add support for log data and trace data, as well as including more RCA models in the benchmark. We welcome and encourage contributions from the open-source community to help us achieve our goals.

\section*{Acknowledgements}
We would like to thank a number of leaders and colleagues from Salesforce who have provided strong support, advice, and contributions to this open-source project. We also would like to extend our thanks to Sai Prasad Mysary, Asharam Yadav, and Sudheen Koyyalummal from Salesforce self-driving database team for our collaboration and partnership in developing early capabilities in AI for database operations that inspired this library. 

\vskip 0.2in
\bibliography{references}

\end{document}